# On-Board Visual Tracking with Unmanned Aircraft System (UAS)

Ashraf Qadir[1], Jeremiah Neubert[2] and William Semke[3]
*University of North Dakota, Grand Forks, North Dakota, 58201*

**This paper presents the development of a real time tracking algorithm that runs on a 1.2 GHz PC/104 computer onboard a small UAV. The algorithm uses zero mean normalized cross correlation to detect and locate an object in the image. A Kalman filter is used to make the tracking algorithm computationally efficient. Object position in an image frame is predicted using the motion model and a search window centered at the predicted position is generated. Object position is updated with the measurement from object detection. The detected position is sent to the motion controller to move the gimbal so that the object stays at the center of the image frame. Detection and tracking is autonomously carried out on the payload computer and the system is able to work in two different methods. The first method starts detecting and tracking using a stored image patch. The second method allows the operator on the ground to select the interest object for the UAV(Unmanned Aerial Vehicle) to track. The system is capable of re-detecting an object, in the event of tracking failure. Performance of the tracking system was verified both in the lab and on the field by mounting the payload on a vehicle and simulating a flight. Tests show that the system can detect and track a diverse set of objects in real time. Flight testing of the system will be conducted at the next available opportunity.**

## I. Introduction

IN recent years small Unmanned Aerial Vehicles (UAV) have generated a lot of interest because of their potential to aid in a wide range of task from law enforcement to precession agriculture. Specifically, visual tracking with small unmanned aircraft systems has applications in surveillance, search and rescue and traffic monitoring etc. They are particularly effective in GPS denied environment for tracking non cooperative targets. For example tracking a car in an urban road or rescuing a human in woods where it is impossible to get GPS position of the interest object. Figure 1 shows an example of UAV application in an urban road.

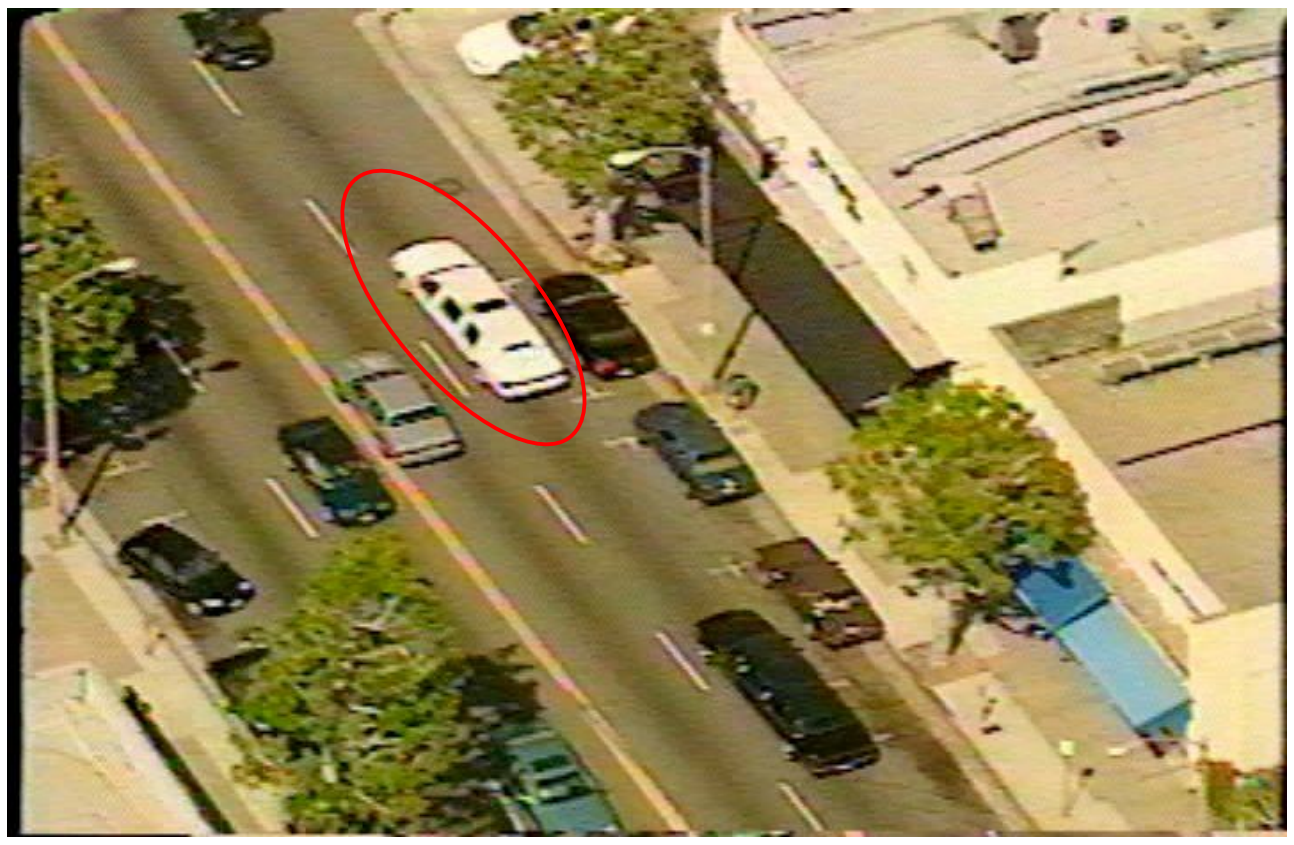

Figure 1. Application of visual tracking with UAV[1]. The UAV will locate and follow the white car in red circle

[1] Graduate Student, Mechanical Engineering, 243 Centennial Drive Stop 8359 Grand Forks ND 58202-8359
[2] Assistant Professor, Mechanical Engineering, 243 Centennial Drive Stop 8359 Grand Forks ND 58202-8359
[3] Associate Professor, Mechanical Engineering, 243 Centennial Drive Stop 8359 Grand Forks ND 58202-8359

A substantial amount of research on visual tracking with small UAV[2-6] has been carried out and a wide range of object recognition and tracking methods have been used. However existing object recognition algorithms are computationally expensive and run on a powerful ground computer. The ground computer receives video information from the UAV and navigation information is then sent to the UAV. A major disadvantage with running the tracking algorithm on a ground computer is its dependency on communication between the ground computer and the UAV. Communication between the UAV and ground computer are often interrupted due to reliability issues with inexpensive commercial off-the-shelf (COTS) equipment, radio frequency interference (RFI) and Electromagnetic interference (EMI) especially when operated near high populated area. Tracking fails when there is a communication failure between the UAV and ground.

The tracking system described in Ref. 2 with small UAVs uses a Commercial off-the-shelf (COTS) image processing software running on a separate ground computer. Video from the onboard camera is transmitted to the ground computer where the images are processed and target information is extracted. Then the information is sent to the guidance and navigation module to calculate the aircraft trajectory. The guidance commands are then sent back to the UAV. This system depends heavily on the communication between the UAV and the ground computer and a communication failure will result in the tracking system failure. Another vision based tracking system for a small UAV with a miniature pan-tilt gimbaled camera was described in Ref. 3. The video sent from the UAV is processed on the ground using a COTS image processing algorithm and then the target position in the camera frame is identified and its position is then sent to the UAV for the UAV/gimbal control. A linear parametrically varying (LPV) filter based motion estimation algorithm has been introduced where the target-loss events due to communication interruption have been modeled as brief instabilities. However the algorithm shows a degradation of performance in presence of target loss events. Ref. 4 also described a vision based tracking system for UAV control. They have implemented their tracking system on a test-bed developed at Universidad Polite`cnica de Madrid. The tracking system used a Scale Invariant Feature Transform (SIFT) algorithm for detecting salient points at every processed frame for visual referencing. Test results show satisfactory matching but at a rate not sufficient for real time tracking and they found tracking speed depends heavily on the size of the search window. An integration of vision measurements along with measurements for the navigation of a *GTMax* unmanned helicopter was proposed in Ref. 6. A geometric active contour method was used to identify contours in a black and white image. However this system was developed to identify building symbols in an urban area assuming the target is a dark square within a much lighter background.

An onboard tracking system eliminates the dependency on the communication with the ground station and makes the system less prone to failure. However small unmanned aircraft systems have limited payload capacity and power budget. As a result an onboard visual tracking system requires a tracking algorithm which is robust and computationally efficient. Having a robust and computationally efficient tracking system poses several challenges. The major challenges arise in the form of target representation, target appearance change, target detection and localization in real time computation. A vehicle on the ground from the UAV flying around 800 hundred feet appears very small. Figure 2 shows the appearance of the object in the image frame captured by the SUNDOG[7, 8] payload camera when flying at around 800 feet. Therefore it is difficult to extract enough local features on the object for feature based tracking. Another challenge is the appearance change of the target object. The target might be subjected to illumination change due to varying lighting condition. A rotation of the object or camera will make the object appear different. A tracking system is needed which is rotation and lighting invariant and capable of tracking small objects. Again detecting the object in each image frame requires searching the image frame to find the best match. If the whole image frame is searched the tracking system becomes computationally expensive. A position prior can reduce the amount of searching and make the algorithm computationally efficient. The gimbal is also moved with the object in order to keep the object at the center of the image frame so that the object does not go out of the camera view.

A real time visual tracking system has been developed to run onboard a small UAV. The target object is represented as a template selected by the operator online or a stored image. The template is matched with the source image frame using zero mean normalized cross correlation to make the tracking system invariant to lighting change. A set of 36 image patch is generated with 10 degrees interval using image warping which represents the set of templates. Now our system becomes rotation invariant. Position of the object is predicted with the motion model using a Kalman filter[9] and a search window centered at the predicted position of the object is generated using the covariance matrix of the Kalman filter. After locating the object with template matching the Kalman filter is updated.

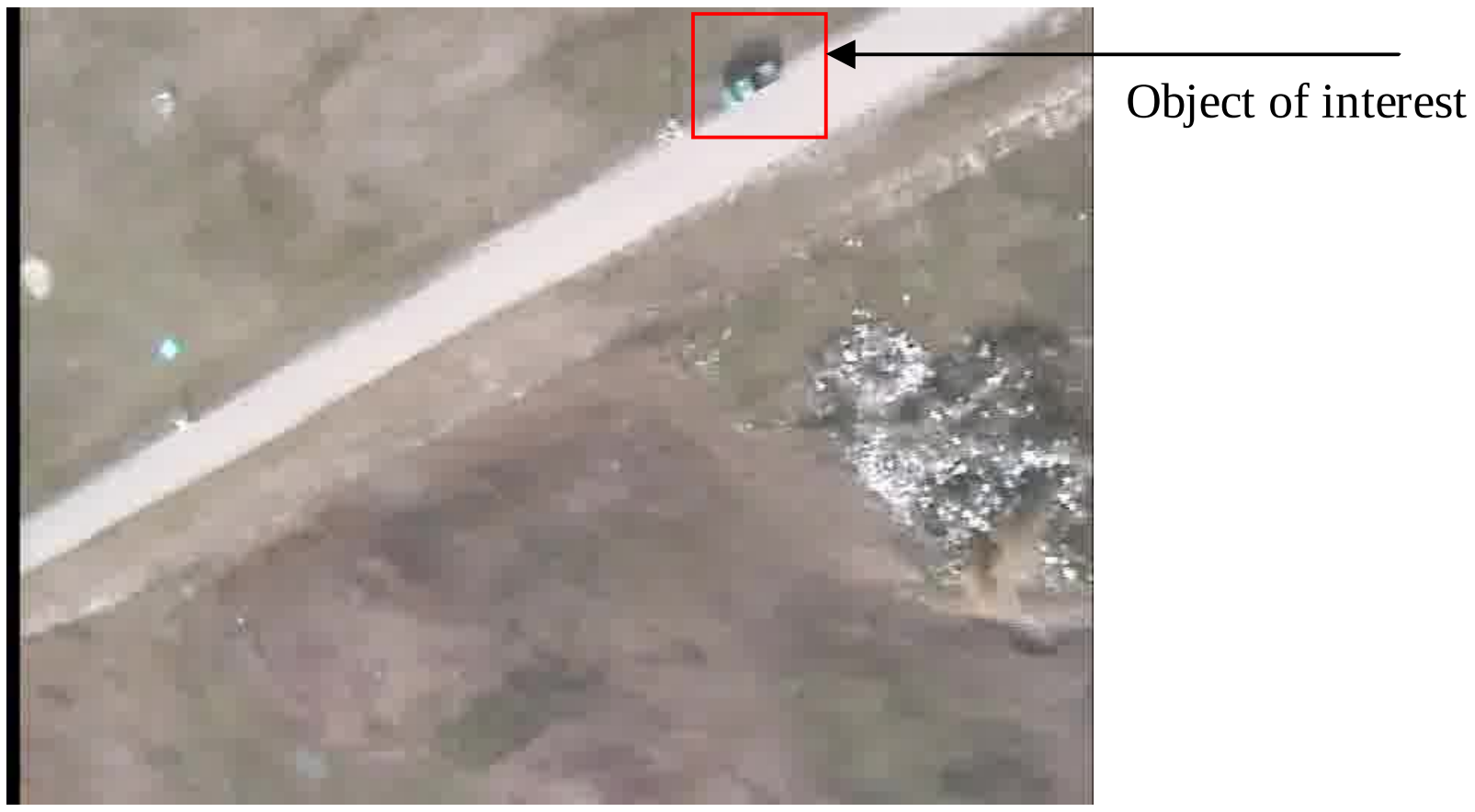

Figure 2. Image captured with the SUNDOG[7,8] payload during previous test flight. The figure shows the appearance of the vehicle inside the red rectangle

The basic architecture of the tracking system is described in Section II. Section III describes the hardware and software implementation of the tracking system. Testing and results of the system are presented in Section IV. Future works with concluding remarks are presented in Section V.

## II. Tracking System Architecture

The basic tracking system developed can be described in brief as:

1. Acquire images
2. Detect the object in an image
3. Maintain the localization of the object in subsequent images through a motion model and
4. Actuate the camera to keep the object of interest centered in the image.

The tracking algorithm is capable of working in two different ways. The first method uses a stored image patch as the template for the tracking system. The patch can be obtained from previous video taken by the UAV. The second method allows the user to specify the object of interest. Samplings of images are sent to the ground station through a 2.4 GHz wireless link using UDP socket. The user then selects the object of interest on the image frame and the image patch is sent to the payload via a pass-through between the autopilot and ground station for flight operation.

### A. Template matching with zero mean normalized cross correlation

The object is detected comparing the template and the image using zero mean normalized cross correlation. The template is slid over the image and correlation coefficient is calculated to detect the position of the template in the image frame. A detailed description of normalized form of cross correlation can be found in Refs. 10 and 11 where they also proposed fast algorithms to calculate zero mean normalized cross correlation coefficient. The normalized correlation coefficient equation is described as

$$C = \frac{\sum \left[ f(x,y) - \bar{f}_{u,v} \right] \left[ t(x-u, y-v) - \bar{t} \right]}{\left\{ \sum \left[ f(x,y) - \bar{f}_{u,v} \right]^2 \left[ t(x-u, y-v) - \bar{t} \right]^2 \right\}^{1/2}}, \tag{1}$$

where $C$ is the Zero Mean Normalized Cross Correlation coefficient, $\bar{t}$ is the mean intensity value of the template and $\bar{f}_{u,v}$ is the mean of the image intensity and $f(x, y)$ in the region under the template. The correlation coefficient value ranges from -1 to +1. A best match between the image region and the template results in a coefficient value of +1 and -1 means a complete disagreement between the template and image. The threshold value for the correlation coefficient in out tracking algorithm is selected as 0.9. Any value equal or above 0.9 is trusted as a true match.

### B. Image Warping

Zero mean normalized cross correlation (ZMNCC) makes the tracking system invariant to intensity changes in the image sequences. However, template matching with ZMNCC detects object where there is only translation or small changes of the interest object shape or orientation. An object viewed from the UAV flying 600/700 feet appears small and its shape does not change much. But both the object and/or camera have rotation when tracking with a UAV. Image warping has been used to make the tracking system rotation invariant. A set of 36 templates have been generated from the original image patch with 10 degrees interval using image warping to accommodate full 360 degree rotation of the object. Templates are then compared with the image to detect the object of interest.

Image warping can be defined as mapping a position $(u,v)$ in the source image to the position $(x,y)$ in the destination image[12]. If a position in the source 2D image expressed in homogeneous coordinates as $x[x, y, 1]^T$, and its corresponding position $x'[u,v,1]^T$ in the destination image, also in homogeneous coordinates, then the mapping can be described as

$$x' = Hx \tag{2}$$

where H denotes the transformation matrix. The transformation can be expressed as

$$\begin{bmatrix} x' \\ y' \\ 1 \end{bmatrix} = \begin{bmatrix} s\cos\alpha & s\sin\alpha & t_x \\ -s\sin\alpha & s\cos\alpha & t_y \\ 0 & 0 & 1 \end{bmatrix} . \begin{bmatrix} x \\ y \\ 1 \end{bmatrix} \tag{3}$$

where $s$ represents scaling and $t(t_x,t_y,1)$ represents translation of the point $x$. For a pure rotation $t_x = t_y = 0$ and $s = 1$.

### C. Kalman Filtering

Moving the template over the entire source image and compute the correlation coefficient at every position is computationally expensive and consumes a lot of time. If the position of the interest object is predicted and only a region around the predicted position is searched to detect the object, the tracking system would become computationally efficient. A Kalman filter[9] has been used to address this issue. The position of the interest object is predicted using a motion model and a search window is generated around the predicted position using the process covariance matrix of the Kalman filter. The position is then estimated using the measurement from the object detection and the process covariance matrix is updated. In the absence of any detection, the state is not updated and search window gets bigger. Two governing equations for extended Kalman filtering[13, 14] are:

$$x_k = f(x_{k-1}, u_{k-1}, w_{k-1}) \tag{4}$$

$$z_k = h(x_k, v_k) \tag{5}$$

Equation (4) is the non linear stochastic difference equation where $x_k$ is the state of the system, $k$ is the time stamp, the function $f$ describes the process model of the system and $w$ is the process noise. Equation (5) relates the measurements $z_k$ to the state $x_k$ of the system at the time stamp $k$.

### D. Actuating the Gimbal

Once detected, the distance between the center of the image and the detected position is computed. The distance is then converted to motor count and sent to the motion controller for the pan and tilt motion of the gimbal. The gimbal actuates accordingly to keep the detected object at the center of image frame.

## III. Implementation

The tracking system was implemented on the SUNDOG (Surveillance by University of North Dakota Observational Gimbal) payload developed by the undergraduate mechanical and electrical students at the University of North Dakota. A customized UAV named “Super Hauler”, owned and operated by the UASE lab, UND houses the SUNDOG payload as well as the Piccolo autopilot[15] with its dedicated control link. The UAV has a wingspan of 144 inches, a wing area of 3680 square inches, and 120 inches of length. The UAV weighs 48 pounds and has 25 pounds of payload carrying capacity.

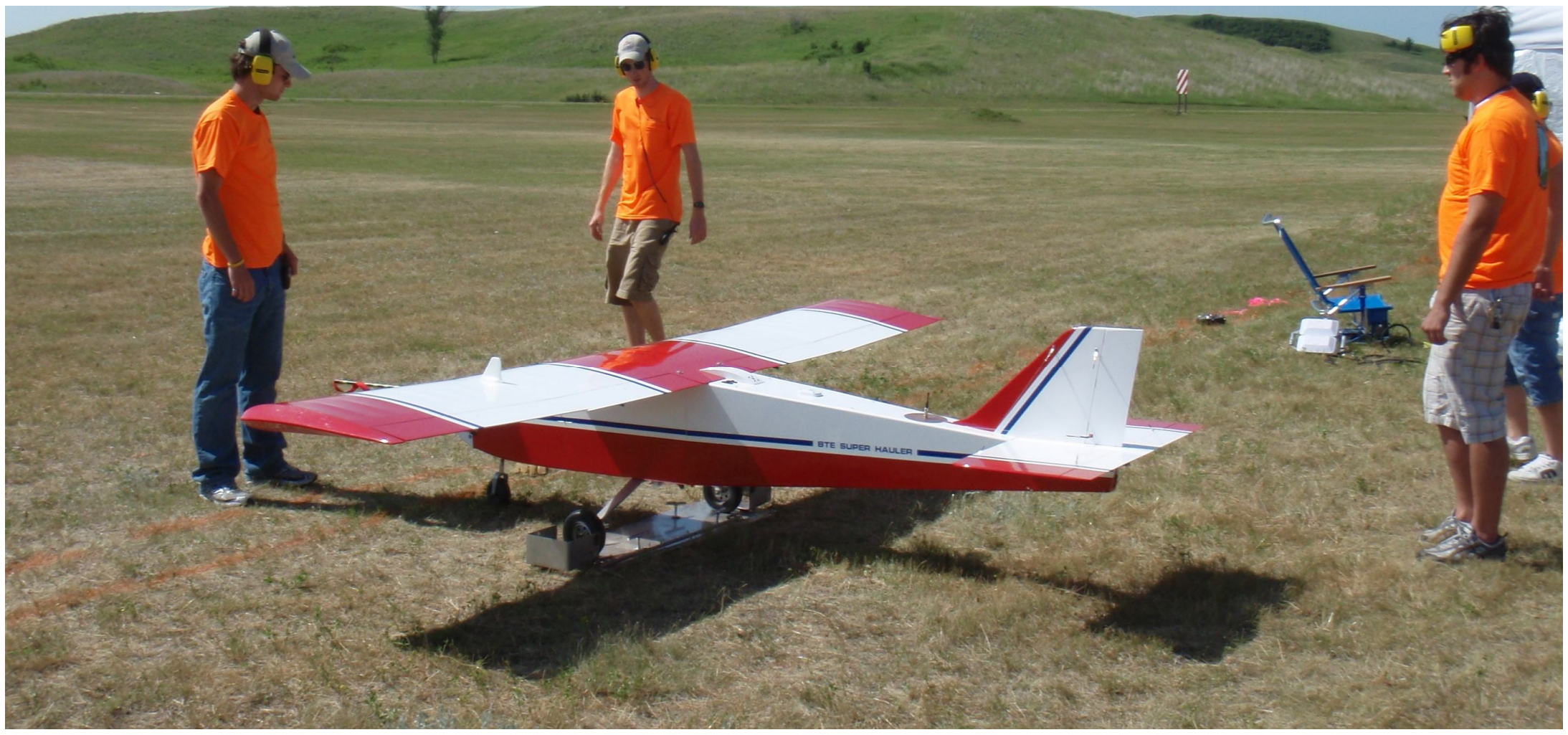

Figure 3. Super Hauler UAV owned and operated by the UND UASE lab

## A. **SUNDOG Payload**

The payload consists of a PC/104+ form factor based computer- essentially a Linux pc on a single printed circuit board (PCB) with frame grabber, additional octal serial port board and wireless card, a three-axis precision pointing system for an Electro-Optical camera and an Infrared camera. A 2.4 GHz PC/104-plus form factor based wireless card and a RTD PC/104- Plus Dual Channel Frame Grabber is stacked with the computer. A color Sony FCBEX980 camera is mounted on the gimbal. The gimbal has 360 degrees rotation and 30 degrees pan and tilt rotation. A PC/104-plus form factor based octal serial port is also stacked with the computer to provide additional serial I/Os for communication with the motion controller, autopilot and other sensors.

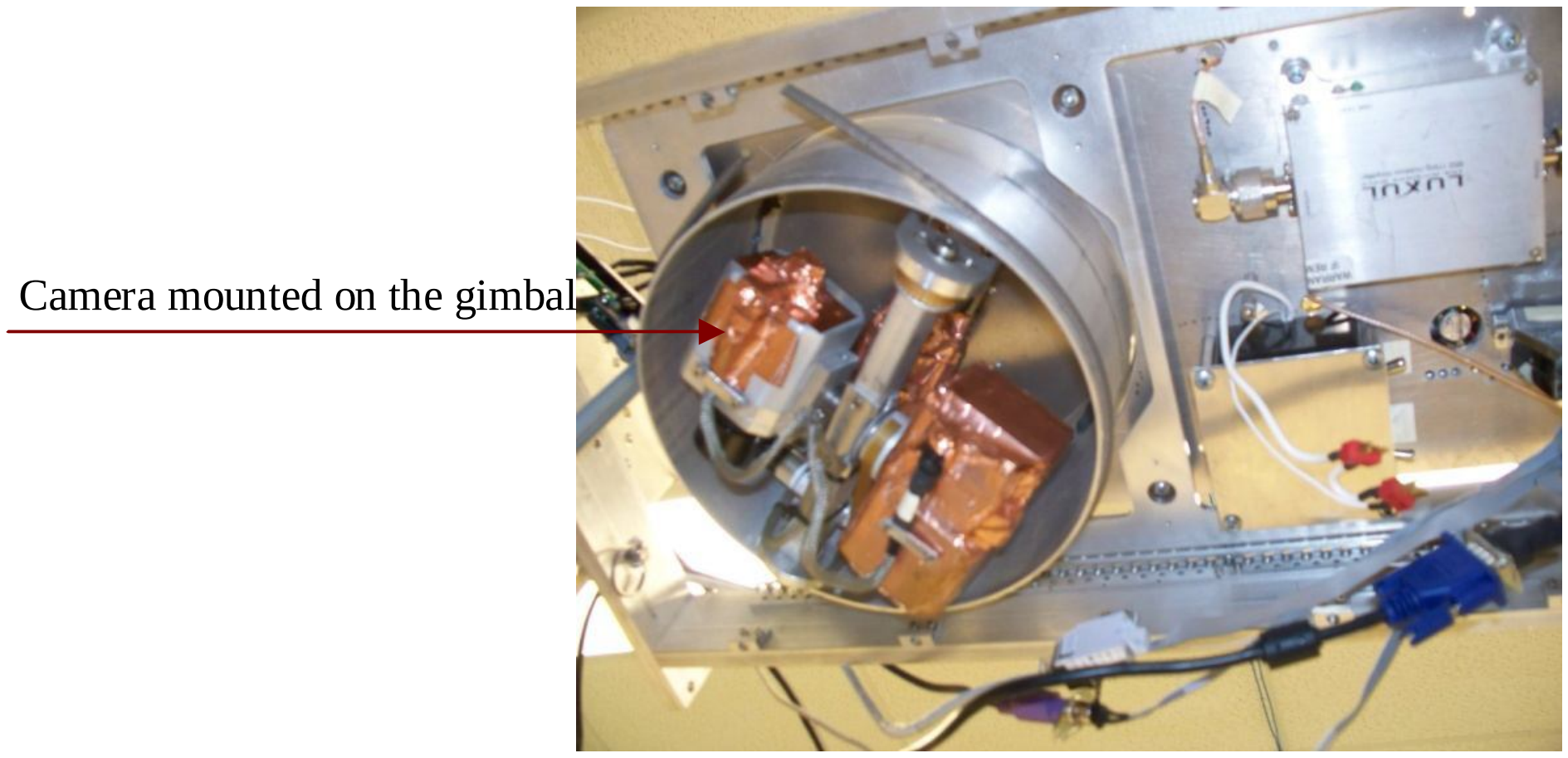

Figure 4. SUNDOG[7,8] payload. Image shows the three-axis gimbal system with the Electro-Optical (EO) and Infrared (IR) cameras mounted on it.

## **B. Motion Controller**

FaulHaber motion controller[16] use pulse-width modulation (PWM) signals to drive the DC servo motors. The drive or amplifier transforms the PWM signal into high amplitude current to turn the motors. They allow for torque control via current regulation. Incremental encoders have been used for position feedback. The controllers are connected via RS232 serial cable to the on-board computer and provide resolution of 100 micro-radians ($0.0057^0$).

**C. Joystic Control**

A joystic control of pointing the gimbal has been implemented for on-line selection of the interest object from the ground. Joystic control allows the ground operator to manually point the gimbal at the target for selection.

**D. Kalman filter implementation**

Kalman filter is used to make the localization process more efficient. The position of the interest object is predicted using a motion model and a search window is defined that is centered at the predicted position. The equations for Kalman filter and their derivation can be found in Refs. 9, 13, and 14. Initializing and implementation of the Kalman filter in our tracking algorithm are described here.

An extended Kalman filter is used for the vision tracking of the complex trajectory (change of acceleration) of the object. The dynamic system was modeled as:

$$\vec{X}_k = e^{\left[\dot{\vec{x}}_{k-1}\Delta t\right]} * \vec{X}_{k-1} \tag{6}$$

The filter is initialized with the following items

1. Process Jacobian: For a 2 degree of freedom system the process Jacobian

$$A = \begin{bmatrix} 1 & 0 & \Delta t & 0 \\ 0 & 1 & 0 & \Delta t \\ 0 & 0 & 1 & 0 \\ 0 & 0 & 0 & 1 \end{bmatrix} \tag{7}$$

Jacobian matrix $A$ is dependent on the time elapsed between observation $k-1$ & $k$ and is denoted $\Delta t$

2. Process noise covariance matrix is initialized as

$$Q = \begin{bmatrix} a_1 & 0 & b_3 & 0 \\ 0 & a_2 & 0 & b_4 \\ b_3 & 0 & \Delta t\sigma_3 & 0 \\ 0 & b_4 & 0 & \Delta t\sigma_4 \end{bmatrix}, \tag{8}$$

where

$$a_i = \Delta t\sigma_i + \frac{1}{3}\Delta t^3\sigma_{i+2}, \tag{9}$$

and

$$b_i = \frac{1}{2}\Delta t^2\sigma_i \tag{10}$$

The exact values of $\sigma$ are not known, reasonable values are selected using experience and knowledge of the process. We have chosen our $\sigma$ as 0.4 after generating the covariance matrix and the search window around the predicted position with different values of sigma during experimentation.

3. Measurement Jacobian

$$H_k = \frac{\partial h}{\partial \vec{X}} = \begin{bmatrix} 1 & 0 & 0 & 0 \\ 0 & 1 & 0 & 0 \end{bmatrix} \tag{11}$$

4. Measurement noise $R_k$ is projected into state space using the equation $V_k R_k V_k^T$. The following measurement noise matrix here has been used:

$$R_k = \begin{bmatrix} 1 & 0 \\ 0 & 1 \end{bmatrix} \tag{12}$$

The steps involved in using Kalman filtering in our vision tracking system are:

1. Initialization ($k$=0): In this stage the whole image is searched for the object due we do not know previously the object position. The object is detected in the image frame and its center is selected as the initial state $\hat{x}_0$ at time $k = 0$. Process covariance matrix $\hat{P}_k$ is also initialized.

2. Prediction ($k$>0)**:** The state of the object $\hat{x}^-_{k=1}$ is predicted using the motion model of the Kalman filter for the next image frame at time $k$=1. This position is considered as the center of the search window to find the object.

3. Correction ($k$>0)**:** In this stage the object is detected within the search window (measurement $z_k$ ) and the state $\hat{x}_k$ and covariance matrix $P_k$ is updated with the measurement data.

Steps 2 and 3 are carried out while the object tracking runs. The size of the search window is dictated by the noise in the prediction and depends on the process covariance matrix. A small process covariance matrix results in a small search window size and implies that the estimation is trusted more. In the absence of detection, there are no measurements and state is not updated. This results in a larger process error covariance and the search window gets bigger.

## IV. Results

### A. Testing with Previous Flight Video

The algorithm was tested with video captured in previous test flights with two objectives in mind. One is to check that our detection algorithm is robust enough to locate objects on the ground and the second one is to tune the algorithm so that it is computationally efficient and capable of tracking real time. An object is selected in one image frame and then tracked in subsequent frames, as shown in the Figure 5. The object is selected in the left image and the template is generated. The purple rectangle represents the selection area in the source image frame. The right image shows the object tracking. The red dot in the image represents the detected object which is the centroid of all matching points between the template and the source image. The algorithm was tested with different threshold values for cross correlation coefficient and it was found that a coefficient value of 0.9 or more gives correct match between the image and template. So a threshold value of 0.9 for the cross correlation coefficient has been selected to use in our tracking algorithm Any value equal to or above 0.9 is accepted as true match. The green rectangle around the object is the search window generated by the covariance matrix of the Kalman filter while the purple rectangle represents the initial position of the object. The search window size depends on the covariance matrix of the Kalman filter. Different search window size was generated by selecting different values of covariance matrix elements ( $\sigma$ ). Then the system was optimized by generating the search window which is big enough to make sure that the object stays inside the search window in successive frames.

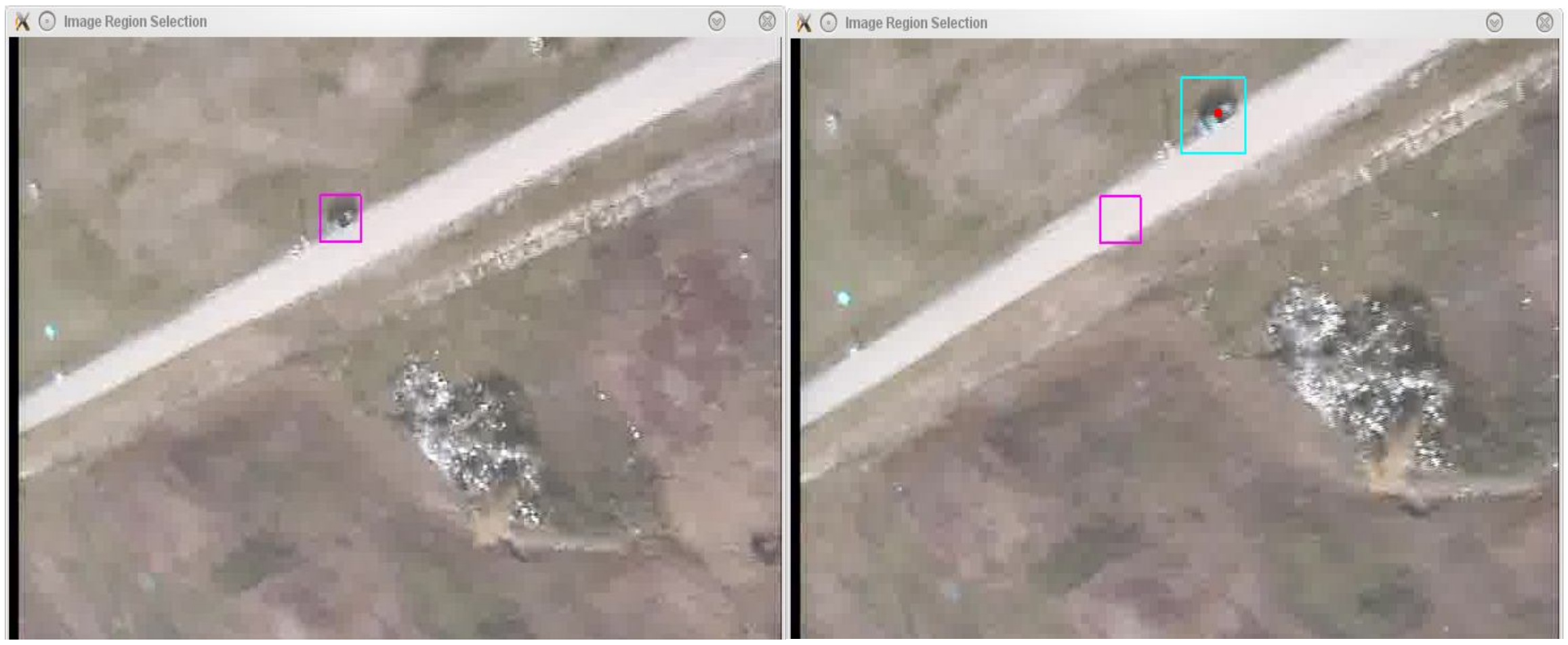

Selecting the object to track　　　　Tracking the object in image sequences

Figure 5: Tracking on the previously captured image frames

### B. Tracking Model Cars

A laboratory experiment has been carried out to verify the performance of the on-board visual tracking system. A small moving car was used as the target object. The payload was mounted on a stand 3 feet (0.9144 meters) above the ground. The goal of the payload was to track the car on the ground and move its gimbal to keep the car at the center of the image frame. The middle section of the top of the car was selected as the template. The size of the template was 22x36 pixels. The tracking algorithm detected the car and started tracking. To verify the robustness and efficiency of the tracking system, the car was translated, rotated and moved around other objects. Tracking was displayed on the screen in real time as the system was running and intermittent image frames were saved for later analysis purpose.

Results of the experiment are shown in Figure 6. The red dot shows the centroid of the detected positions and the aqua rectangle around it is the search window generated by the covariance matrix of the Kalman filter. The object appeared with different position and orientation as it was moved and the tracking system was able to detect the car. The gimbal was actuated to keep the car at the center of the image frame. The tracking system exhibited its ability to real time track and re-detect objects in the event of tracking failure in one frame.

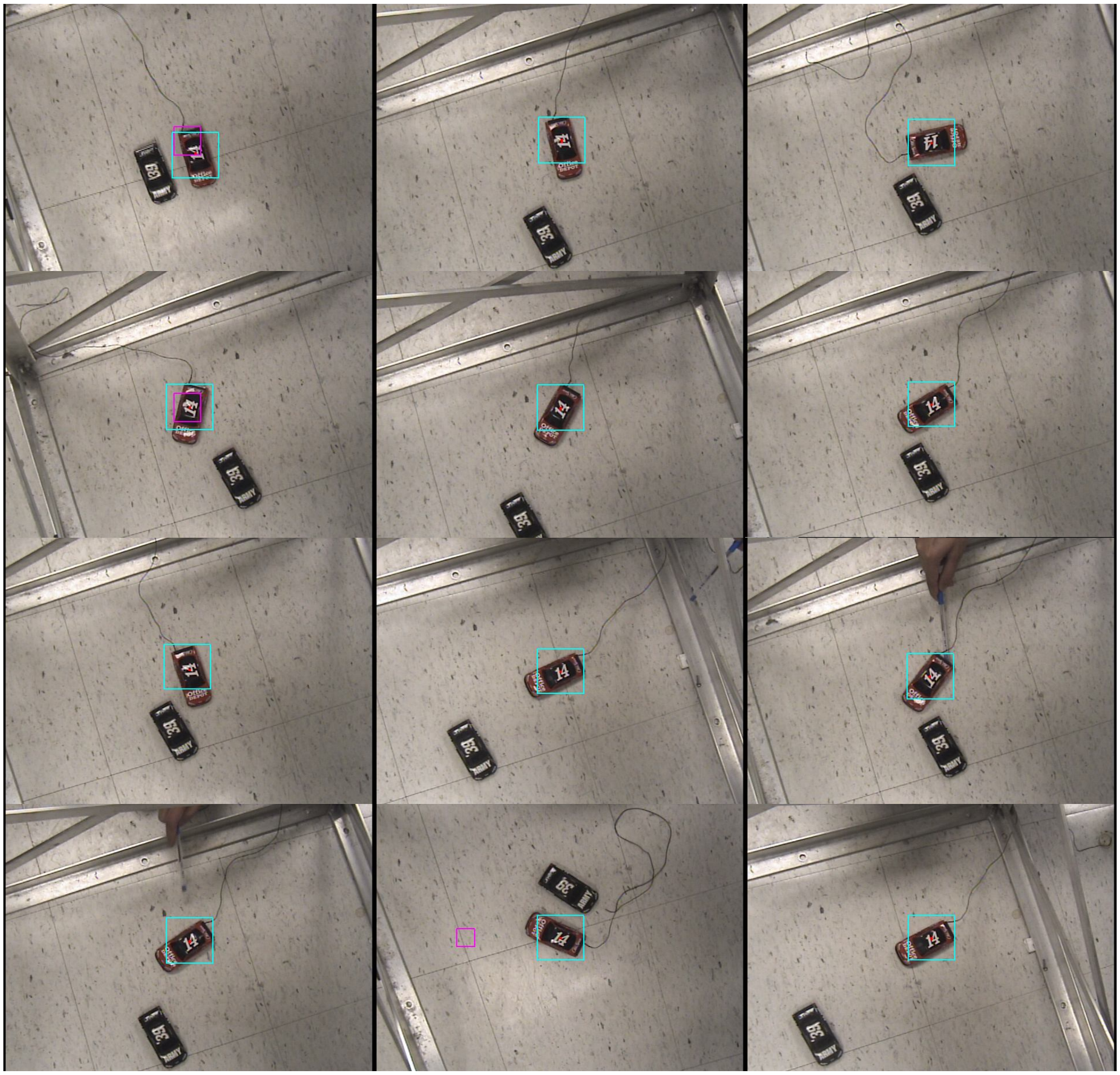

Figure 6. Tracking model car in the lab. The car was translated and rotated to test the performance of the tracking algorithm

Online selection of an object of interest and sending this data to the payload was also tested in the lab using a direct serial communication between ground computer and the payload. Video frames were sent over the ethernet from the payload to the computer used as a ground station. The car was selected on the ground computer screen and the image patch was sent to the payload using the RS232 communication. Upon receiving the image patch, the payload started tracking.

### C. Mobile Ground Test

Lastly, a mobile ground test of the system has been carried out to by simulating flight environment for the UAV to test the tracking system's performance where the plane with the payload is mounted on a truck. The ground station was setup in the middle of a parking lot and the truck carrying the plane was driven around. A simulated flight environment has been created in the autopilot software. The payload transmits image frames to the ground computer using the wireless network. An object was selected on the ground computer and then sent to the payload via piccolo payload pass-through. The object was then being tracked while the truck is moving. The test-bed for the experiment is shown in Figure 7.

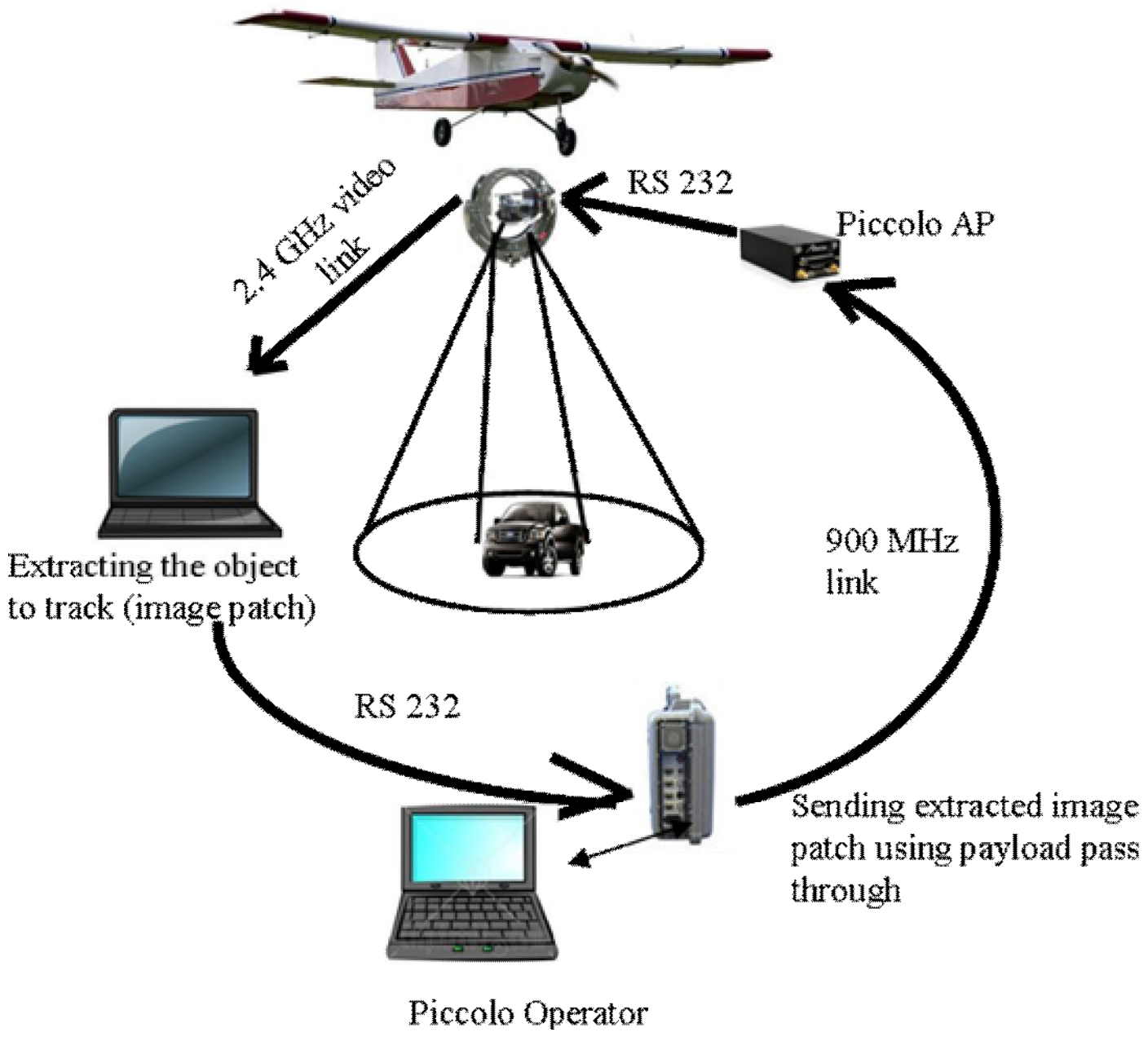

Figure 7. Test-bed for mobile ground test

Results of the mobile ground tests are shown in Figure 8. The truck was driven at 15 miles per hour with the payload mounted at the back and the selected object for tracking was around 270 feet away from the payload. Test results show the tracking systems ability to track an object when mounted on a moving vehicle and re-detecting the object in the event of tracking failure in one image frame. The purple rectangle represents the online selection of the object of interest in the image frame. Once detected, the gimbal was actuated to move the object of interest at the center of the image frame. The green rectangle around the object is the search window generated by the covariance matrix of the Kalman filter. The red dot on the object is the centroid of all matched points.

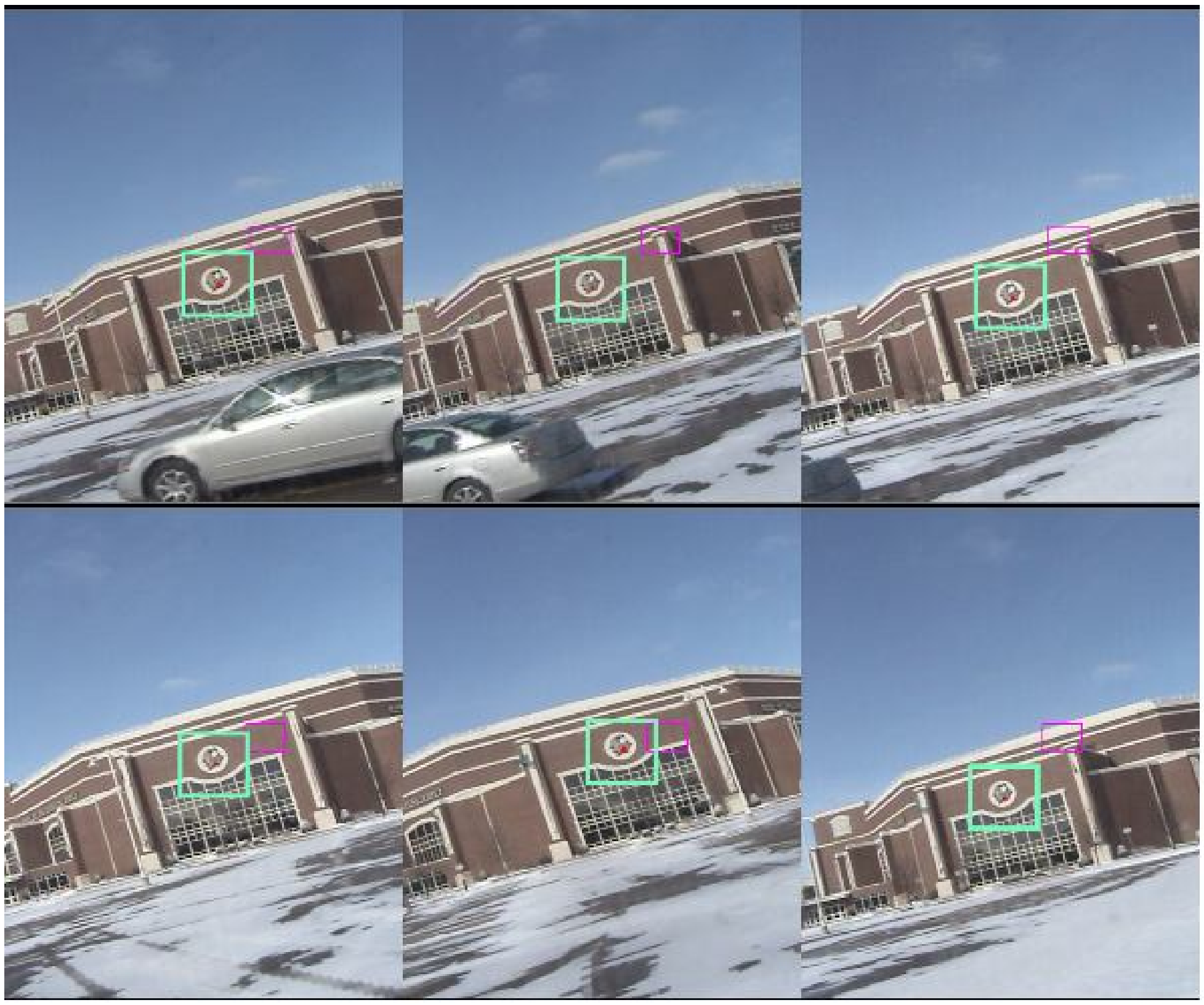

Figure 8. Tracking with the payload mounted on a truck for mobile ground test

## V. Conclusion and Future Work

In this paper an implementation of on-board visual tracking with small UAV has been presented. The tracking system locates the object of interest using zero mean normalized cross correlation between object template and source image. The system was implemented on the SUNDOG payload and was tested using several experimentations including a full hardware in the loop test in the lab and a mobile ground test. The tracking algorithm has the ability to re-detect and track in the event of loss of tracking. In hardware in the loop test the tracking system shows continuous uninterrupted tracking in real time. During mobile ground test the tracking algorithm failed when there is a start or stop of the vehicle carrying the payload. This was due to the vibration during the starting and stopping of the truck causing the images to be blurred. The system quickly recovered from this failure by expanding the search area. The tracking system is invariant to changes in illumination or rotation of the object. The system assumes that large scale changes do not occur. Given previous flight videos this assumption appears valid. A simple linear projective transformation can be used in the tracking system with the aircraft altitude information to accommodate scaling. Better control of the camera parameters such as gain, and exposure will reduce the blurriness in the images caused by vibration or very fast movement of both the camera and the object.

Actual flight test of the tracking system will be conducted in the next available dates. A progressive scan camera can reduce the flickering effect while tracking fast moving object. Controlling the brightness and exposure time will eliminate blurriness in the image frames.

## V. Acknowledgement

This research was supported in part by Department of Defense contract number FA4861-06-C-C006, “Unmanned Aerial System Remote Sense and Avoid System and Advanced Payload Analysis and Investigation,” and North Dakota Department of Commerce grant entitled ”UND Center of Excellence for UAV and Simulation

Applications." The authors appreciate the contributions of the Unmanned Aircraft Systems Engineering (UASE) Laboratory team at the University of North Dakota.